\documentclass{article}

\usepackage[preprint]{neurips_2026}
\usepackage[utf8]{inputenc} 
\usepackage[T1]{fontenc}    
\usepackage{hyperref}       
\usepackage{url}            
\usepackage{booktabs}       
\usepackage{amsfonts}       
\usepackage{nicefrac}       
\usepackage{microtype}      
\usepackage{xcolor}         

\usepackage{acronym}
\usepackage{subcaption}
\usepackage{graphicx}
\usepackage{array}

\newcommand{\blfootnote}[1]{%
  \begingroup
  \renewcommand{\thefootnote}{}\footnote{#1}%
  \addtocounter{footnote}{-1}%
  \endgroup
}

\acrodef{ADC}[ADC]{Analog-to-Digital Converter}
\acrodef{ADEXP}[AdExp-IF]{Adaptive Exponential Integrate-and-Fire}
\acrodef{ADM}[ADM]{Asynchronous Delta Modulator}
\acrodef{AE}[AE]{Address-Event}
\acrodef{AER}[AER]{Address-Event Representation}
\acrodef{AEX}[AEX]{AER EXtension board}
\acrodef{AFE}[AFE]{Analog Front-End}
\acrodef{AFM}[AFM]{Atomic Force Microscope}
\acrodef{AGC}[AGC]{Automatic Gain Control}
\acrodef{AI}[AI]{Artificial Intelligence}
\acrodef{AMDA}[AMDA]{AER Motherboard with D/A converters}
\acrodef{AMPA}[AMPA]{$\alpha$-Amino-3-hydroxy-5-methyl-4-isoxazolepropionic Acid}
\acrodef{ANN}[ANN]{Artificial Neural Network}
\acrodef{API}[API]{Application Programming Interface}
\acrodef{APMOM}[APMOM]{Alternate Polarity Metal On Metal}
\acrodef{APS}[APS]{Active Pixel Sensor}
\acrodef{ARM}[ARM]{Advanced RISC Machine}
\acrodef{ASIC}[ASIC]{Application Specific Integrated Circuit}
\acrodef{ATIS}[ATIS]{Asynchronous Temporal Imaging Sensor}
\acrodef{BCM}[BMC]{Bienenstock-Cooper-Munro}
\acrodef{BD}[BD]{Bundled Data}
\acrodef{BEOL}[BEOL]{Back-end of Line}
\acrodef{BG}[BG]{Bias Generator}
\acrodef{BMI}[BMI]{Brain-Machince Interface}
\acrodef{BTB}[BTB]{Band-to-Band tunnelling}
\acrodef{BU}[BU]{Bottom-Up}
\acrodef{CA}[CA]{Cortical Automaton}
\acrodef{CAD}[CAD]{Computer Aided Design}
\acrodef{CAM}[CAM]{Content Addressable Memory}
\acrodef{CAR-FAC}[CAR-FAC]{Cascade of Asymmetric Resonators with Fast-Acting Compression}
\acrodef{CAVIAR}[CAVIAR]{Convolution AER Vision Architecture for Real-Time}
\acrodef{CCN}[CCN]{Cooperative and Competitive Network}
\acrodef{CDR}[CDR]{Clock-Data Recovery}
\acrodef{CFC}[CFC]{Current to Frequency Converter}
\acrodef{CHP}[CHP]{Communicating Hardware Processes}
\acrodef{CMIM}[CMIM]{Metal-Insulator-Metal Capacitor}
\acrodef{CML}[CML]{Current Mode Logic}
\acrodef{CMOL}[CMOL]{Hybrid CMOS nanoelectronic circuits}
\acrodef{CMOS}[CMOS]{Complementary Metal-Oxide-Semiconductor}
\acrodef{CNN}[CNN]{Convolutional Neural Network}
\acrodef{CNS}[CNS]{central Nervous System}
\acrodef{COTS}[COTS]{Commercial Off-The-Shelf}
\acrodef{CPG}[CPG]{Central Pattern Generator}
\acrodef{CPLD}[CPLD]{Complex Programmable Logic Device}
\acrodef{CPU}[CPU]{Central Processing Unit}
\acrodef{CSM}[CSM]{Cortical State Machine}
\acrodef{CSP}[CSP]{Constraint Satisfaction Problem}
\acrodef{CTXCTL}[CTXCTL]{CortexControl}
\acrodef{CV}[CV]{Coefficient of Variation}
\acrodef{DAC}[DAC]{Digital to Analog Converter}
\acrodef{DAS}[DAS]{Dynamic Auditory Sensor}
\acrodef{DAVIS}[DAVIS]{Dynamic and Active Pixel Vision Sensor}
\acrodef{DBN}[DBN]{Deep Belief Network}
\acrodef{DBS}[DBS]{Deep Brain Stimulation}
\acrodef{DEA}[DEA]{Dielectric elastomer actuator}
\acrodef{DFA}[DFA]{Deterministic Finite Automaton}
\acrodef{DIBL}[DIBL]{Drain-Induced Barrier-Lowering}
\acrodef{DI}[DI]{Delay Insensitive}
\acrodef{divmod3}[DIVMOD3]{Divisibility of a number by three}
\acrodef{DMA}[DMA]{Direct Memory Access}
\acrodef{DNF}[DNF]{Dynamic Neural Field}
\acrodef{DNN}[DNN]{Deep Neural Network}
\acrodef{DoF}[DoF]{Degrees of Freedom}
\acrodef{DPE}[DPE]{Dynamic Parameter Estimation}
\acrodef{DSP}[DSP]{Digital Signal Processor}
\acrodef{DPI}[DPI]{Differential Pair Integrator}
\acrodef{DRAM}[DRAM]{Dynamic Random Access Memory}
\acrodef{DR}[DR]{Dual Rail}
\acrodef{DRRZ}[DR-RZ]{Dual-Rail Return-to-Zero}
\acrodef{DSP}[DSP]{Digital Signal Processor}
\acrodef{DVS}[DVS]{Dynamic Vision Sensor}
\acrodef{DYNAP}[DYNAP]{Dynamic Neuromorphic Asynchronous Processor}
\acrodef{EBL}[EBL]{Electron Beam Lithography}
\acrodef{ECG}[ECG]{Electrocardiography}
\acrodef{ECoG}[ECoG]{Electrocorticography}
\acrodef{EDC}[EDC]{Event--Driven Camera}
\acrodef{EDVAC}[EDVAC]{Electronic Discrete Variable Automatic Computer}
\acrodef{EEG}[EEG]{Electroencephalography}
\acrodef{EIN}[EIN]{Excitatory-Inhibitory Network}
\acrodef{EM}[EM]{Expectation Maximization}
\acrodef{EMG}[EMG]{Electromyography}
\acrodef{EOG}[EOG]{Electrooculogram}
\acrodef{EPSC}[EPSC]{Excitatory Post-Synaptic Current}
\acrodef{EPSP}[EPSP]{Excitatory Post-Synaptic Potential}
\acrodef{EZ}[EZ]{Epileptogenic Zone}
\acrodef{FDSOI}[FDSOI]{Fully-Depleted Silicon on Insulator}
\acrodef{FET}[FET]{Field-Effect Transistor}
\acrodef{FFT}[FFT]{Fast Fourier Transform}
\acrodef{FI}[F-I]{Frequency--Current}
\acrodef{FMA}[FMA]{Floating Microelectrode Array} 
\acrodef{FNN}[FNN]{Feed-forward Neural Network}
\acrodef{FPGA}[FPGA]{Field Programmable Gate Array}
\acrodef{FR}[FR]{Fast Ripple}
\acrodef{FSA}[FSA]{Finite State Automaton}
\acrodef{FSM}[FSM]{Finite State Machine}
\acrodef{GABA}[GABA]{$\gamma$-Aminobutanoic Acid}
\acrodef{GIDL}[GIDL]{Gate-Induced Drain Leakage}
\acrodef{GOPS}[GOPS]{Giga-Operations per Second}
\acrodef{GPIO}[GPIO]{General Purpose I/O}
\acrodef{GPU}[GPU]{Graphical Processing Unit}
\acrodef{GT}[GT]{Ground Truth}
\acrodef{GUI}[GUI]{Graphical User Interface}
\acrodef{HAL}[HAL]{Hardware Abstraction Layer}
\acrodef{HFO}[HFO]{High Frequency Oscillation}
\acrodef{HH}[H\&H]{Hodgkin \& Huxley}
\acrodef{HMM}[HMM]{Hidden Markov Model}
\acrodef{HR}[HR]{Human Readable}
\acrodef{HRS}[HRS]{High-Resistive State}
\acrodef{HSE}[HSE]{Handshaking Expansion}
\acrodef{HW}[HW]{Hardware}
\acrodef{hWTA}[hWTA]{Hard Winner-Take-All}
\acrodef{IC}[IC]{Integrated Circuit}
\acrodef{ICA}[ICA]{Indipendent Component Analysis}
\acrodef{ICT}[ICT]{Information and Communication Technology}
\acrodef{iEEG}[iEEG]{Intracranial Electroencephalography}
\acrodef{IF2DWTA}[IF2DWTA]{Integrate \& Fire 2-Dimensional WTA}
\acrodef{IF}[I\&F]{Integrate-and-Fire}
\acrodef{IFSLWTA}[IFSLWTA]{Integrate \& Fire Stop Learning WTA}
\acrodef{IMU}[IMU]{Inertial Measurement Unit}
\acrodef{INCF}[INCF]{International Neuroinformatics Coordinating Facility}
\acrodef{INI}[INI]{Institute of Neuroinformatics}
\acrodef{IO}[I/O]{Input/Output}
\acrodef{IoT}[IoT]{Internet of Things}
\acrodef{IoU}[IoU]{Intersection over Union}
\acrodef{IP}[IP]{Intellectual Property}
\acrodef{IPSC}[IPSC]{Inhibitory Post-Synaptic Current}
\acrodef{IPSP}[IPSP]{Inhibitory Post-Synaptic Potential}
\acrodef{ISI}[ISI]{Inter-Spike Interval}
\acrodef{JFLAP}[JFLAP]{Java - Formal Languages and Automata Package}
\acrodef{LCE}[LCE]{liquid crystal elastomer} 
\acrodef{LEDR}[LEDR]{Level-Encoded Dual-Rail}
\acrodef{LFP}[LFP]{Local Field Potential}
\acrodef{LIFE}[LIFE]{Longitudinal Intrafascicular Electrodes}
\acrodef{LIF}[LI\&F]{Leaky Integrate-and-Fire}
\acrodef{LLC}[LLC]{Low Leakage Cell}
\acrodef{LMS}[LMS]{Least Mean Squares}
\acrodef{LNA}[LNA]{Low-Noise Amplifier}
\acrodef{LPF}[LPF]{Low Pass Filter}
\acrodef{LRS}[LRS]{Low-Resistive State}
\acrodef{LSM}[LSM]{Liquid State Machine}
\acrodef{LTD}[LTD]{Long Term Depression}
\acrodef{LTI}[LTI]{Linear Time-Invariant}
\acrodef{LTP}[LTP]{Long Term Potentiation}
\acrodef{LTU}[LTU]{Linear Threshold Unit}
\acrodef{LUT}[LUT]{Look-Up Table}
\acrodef{LVDS}[LVDS]{Low Voltage Differential Signaling}
\acrodef{MCMC}[MCMC]{Markov-Chain Monte Carlo}
\acrodef{MEA}[MEA]{Multielectrode Arrays}
\acrodef{MEMS}[MEMS]{Micro Electro Mechanical System}
\acrodef{MFR}[MFR]{Mean Firing Rate}
\acrodef{MIM}[MIM]{Metal Insulator Metal}
\acrodef{ML}[ML]{Machine Learning}
\acrodef{MLP}[MLP]{Multilayer Perceptron}
\acrodef{MOSCAP}[MOSCAP]{Metal Oxide Semiconductor Capacitor}
\acrodef{MOSFET}[MOSFET]{Metal Oxide Semiconductor Field-Effect Transistor}
\acrodef{MOS}[MOS]{Metal Oxide Semiconductor}
\acrodef{MRI}[MRI]{Magnetic Resonance Imaging}
\acrodef{NAS}[NAS]{Neuromorphic Auditory Sensor}
\acrodef{NCS}[NCS]{Neuromorphic Cognitive Systems}
\acrodef{NDFSM}[NDFSM]{Non-deterministic Finite State Machine} 
\acrodef{ND}[ND]{Noise-Driven}
\acrodef{NEF}[NEF]{Neural Engineering Framework}
\acrodef{NHML}[NHML]{Neuromorphic Hardware Mark-up Language}
\acrodef{NIL}[NIL]{Nano-Imprint Lithography}
\acrodef{NI}[NI]{Neural Interface}
\acrodef{NMDA}[NMDA]{\textit{N}-Methyl-\textsc{d}-aspartate}
\acrodef{NME}[NE]{Neuromorphic Engineering}
\acrodef{NN}[NN]{Neural Network}
\acrodef{NOC}[NoC]{Network-on-Chip}
\acrodef{NRZ}[NRZ]{Non-Return-to-Zero}
\acrodef{NSM}[NSM]{Neural State Machine}
\acrodef{OR}[OR]{Operating Room}
\acrodef{OTA}[OTA]{Operational Transconductance Amplifier}
\acrodef{PCB}[PCB]{Printed Circuit Board}
\acrodef{PCHB}[PCHB]{Pre-Charge Half-Buffer}
\acrodef{PCM}[PCM]{Phase Change Memory}
\acrodef{PC}[PC]{Personal Computer}
\acrodef{PDK}[PDK]{Process Design Kit}
\acrodef{PE}[PE]{Phase Encoding}
\acrodef{PFA}[PFA]{Probabilistic Finite Automaton}
\acrodef{PFC}[PFC]{Prefrontal Cortex}
\acrodef{PFM}[PFM]{Pulse Frequency Modulation}
\acrodef{PGA}[PGA]{Programmable Gain Amplifier}
\acrodef{PID}[PID]{Proportional Integral Derivative}
\acrodef{PNI}[PNI]{Peripheral Nerve Interface}
\acrodef{PNS}[PNS]{Peripheral Nervous System}
\acrodef{PPG}[PPG]{Photoplethysmography}
\acrodef{PR}[PR]{Production Rule}
\acrodef{PSC}[PSC]{Post-Synaptic Current}
\acrodef{PSP}[PSP]{Post-Synaptic Potential}
\acrodef{PSTH}[PSTH]{Peri-Stimulus Time Histogram}
\acrodef{PV}[PV]{Parvalbumin}
\acrodef{PWM}[PWM]{Pulse--Width Modulation}
\acrodef{QDI}[QDI]{Quasi Delay Insensitive}
\acrodef{RAM}[RAM]{Random Access Memory}
\acrodef{RA}[RA]{Resected Area}
\acrodef{RF}[RF]{Receptive Field}
\acrodef{RDF}[RDF]{Random Dopant Fluctuation}
\acrodef{RELU}[ReLu]{Rectified Linear Unit}
\acrodef{RLS}[RLS]{Recursive Least-Squares}
\acrodef{RMSE}[RMSE]{Root Mean Square-Error}
\acrodef{RRMSE}[RMSE]{Relative Root Mean Square Error}
\acrodef{RMS}[RMS]{Root Mean Square}
\acrodef{RNN}[RNN]{Recurrent Neural Network}
\acrodef{ROI}[ROI]{Region of Interest}
\acrodef{ROLLS}[ROLLS]{Reconfigurable On-Line Learning Spiking}
\acrodef{RRAM}[R-RAM]{Resistive Random Access Memory}
\acrodef{R}[R]{Ripple}
\acrodef{RISC}[RISC]{Reduced Instruction Set Computer}
\acrodef{ROS}[ROS]{Robot Operating System}
\acrodef{RSA}[RSA]{Respiratory Sinus Arrhythmia}
\acrodef{SAC}[SAC]{Selective Attention Chip}
\acrodef{SAT}[SAT]{speed-accuracy trade-off}
\acrodef{SCI}[SCI]{Spinal Cord Injury}
\acrodef{SCX}[SCX]{Silicon CorteX}
\acrodef{SD}[SD]{Signal-Driven}
\acrodef{SEM}[SEM]{Spike-based Expectation Maximization}
\acrodef{SFA}[SFA]{Spike frequency adaptation}
\acrodef{SMA}[SMA]{Shape Memory Alloy} 
\acrodef{SLAM}[SLAM]{Simultaneous Localization and Mapping}
\acrodef{SMA}[SMA]{Shape Memory Alloy}
\acrodef{SMCT}[SMCT]{Sensory Motor Contingency Theory} 
\acrodef{SNN}[SNN]{Spiking Neural Network}
\acrodef{SNR}[SNR]{Signal to Noise Ratio}
\acrodef{SOC}[SoC]{System-On-Chip}
\acrodef{SOI}[SOI]{Silicon on Insulator}
\acrodef{SOZ}[SOZ]{Seizure Onset Zone}
\acrodef{SP}[SP]{Separation Property}
\acrodef{SPI}[SPI]{Serial Peripheral Interface}
\acrodef{SRAM}[SRAM]{Static Random Access Memory}
\acrodef{SST}[SST]{Somatostatin}
\acrodef{STDP}[STDP]{Spike-Timing Dependent Plasticity}
\acrodef{STD}[STD]{Short-Term Depression}
\acrodef{STF}[STF]{Short-term Facilitation}
\acrodef{STP}[STP]{Short-Term Plasticity}
\acrodef{STT-MRAM}[STT-MRAM]{Spin-Transfer Torque Magnetic Random Access Memory}
\acrodef{STT}[STT]{Spin-Transfer Torque}
\acrodef{SVM}[SVM]{Support Vector Machine}
\acrodef{SW}[SW]{Software}
\acrodef{sWTA}[sWTA]{soft Winner-Take-All}
\acrodef{TCAM}[TCAM]{Ternary Content-Addressable Memory}
\acrodef{TD}[TD]{Top-Down}
\acrodef{TFT}[TFT]{Thin Film Transistor}
\acrodef{TIME}[TIME]{Transverse Intrafascicular Multichannel Electrode}
\acrodef{TLE}[TLE]{Temporal Lobe Epilepsy}
\acrodef{UEA}[UEA]{Utah Electrode Array}
\acrodef{USB}[USB]{Universal Serial Bus}
\acrodef{USEA}[USEA]{Utah Slanted Electrode Array}
\acrodef{VHDL}[VHDL]{VHSIC Hardware Description Language}
\acrodef{VHSIC}[VHSIC]{Very High Speed Integrated Circuits}
\acrodef{VIP}[VIP]{Vasoactive Intestinal Peptide}
\acrodef{VLSI}[VLSI]{Very Large Scale Integration}
\acrodef{VNS}[VNS]{Vagal Nerve Stimulation}
\acrodef{VM}[VM]{Von Mises}
\acrodef{VOR}[VOR]{Vestibulo-Ocular Reflex}
\acrodef{VSA}[VSA]{Vector Symbolic Architecture}
\acrodef{WCST}[WCST]{Wisconsin Card Sorting Test}
\acrodef{WTA}[WTA]{Winner-Take-All}
\acrodef{XML}[XML]{eXtensible Mark-up Language}
\acrodef{YARP}[YARP]{Yet Another Robot Platform}

\title{Event-based Selective Attention for Multi-resolution Fast \ac{ROI} Detection}

\author{
  \begin{tabular}{cc}
    \begin{minipage}[t]{0.45\textwidth}
      \centering
      Luca Peres$^*$ \\
      \normalfont International Centre for Neuromorphic Systems (ICNS) \\
      The University of Manchester \\
      Manchester, United Kingdom \\
      \texttt{luca.peres-2@manchester.ac.uk}
    \end{minipage}
    &
    \begin{minipage}[t]{0.45\textwidth}
      \centering
      Giulia D'Angelo$^*$ \\
      \normalfont Department of Cybernetics \\
      Faculty of Electrical Engineering \\
      Czech Technical University in Prague \\
      Prague, Czech Republic \\
      \texttt{giulia.dangelo@fel.cvut.cz}
    \end{minipage}
    \\
    \noalign{\vskip 2em}
    \begin{minipage}[t]{0.45\textwidth}
      \centering
      Chiara Bartolozzi \\
      \normalfont Istituto Italiano di Tecnologia \\
      Genoa, Italy \\
      \phantom{x} \\
      \texttt{chiara.bartolozzi@iit.it}
    \end{minipage}
    &
    \begin{minipage}[t]{0.45\textwidth}
      \centering
      Oliver Rhodes \\
      \normalfont International Centre for Neuromorphic Systems (ICNS)\\
      The University of Manchester \\
      Manchester, United Kingdom \\
      \texttt{oliver.rhodes@manchester.ac.uk}
    \end{minipage}
  \end{tabular}
}

\begin{document}

\maketitle
\begingroup
\hypersetup{hidelinks}
\blfootnote{$^*$Corresponding author.}
\endgroup

\begin{abstract}
Neuromorphic vision systems operate under strict constraints on bandwidth, memory, and energy, particularly at the edge, motivating early mechanisms for data reduction and selective processing. In this work, we investigate a multi-scale training-free, saliency-based, bottom-up visual attention model that operates directly on low-resolution event-based input and selects Regions of Interest (\acp{ROI}) from the visual scene. The model is evaluated across multiple downscaling factors applied to the incoming event stream, with input resolutions reduced by up to 256$\times$ relative to full resolution. Performance is assessed on the Prophesee Automotive dataset, the largest publicly available event-based dataset, demonstrating robust ROI selection across different scales on a real-world use-case. The proposed approach is capable of detecting \acp{ROI} belonging to multiple object classes, including various vehicle types, pedestrians, traffic lights, and traffic signs, with accuracy up to 70.8\%, while operating at millisecond temporal resolution, 16$\times$ finer than the temporal resolution provided by the dataset ground truth. These results highlight the potential of combining early event downscaling with saliency-based attention as an effective front-end for efficient edge neuromorphic vision systems.
\end{abstract}

\section{Introduction}
\label{sec:introduction}

The human visual system is constantly stimulated by an overwhelming stream of data from the surrounding environment. Mechanisms of selective visual attention~\cite{rizzolatti1983mechanisms} address this challenge by significantly reducing the amount of data to be processed, directing focus toward specific Regions of Interest (\acp{ROI}). Visual attention entails serial focalization over a sequence of specific targets in the visual scene~\cite{james1890principles} to ideally interact with an unconstrained and dynamic environment. 
Agents immersed in complex environments can take advantage of such mechanisms efficiently, organising their sensory inputs to perceive and explore their surroundings proficiently. 

In biological systems, selective attention is coupled with foveation, whereby the high resolution region of the retina is shifted to focus on the ROI~\cite{bandera1989foveal}. In current artificial vision systems, where the sensor has uniform resolution, one, or multiple \acp{ROI} can be selected and processed.
This approach lowers data complexity and cost, enabling adaptive, energy-efficient vision for real-time applications~\cite{yeasin2005foveated}, without compromising accuracy~\cite{lubana2018digital}. 
Additionally, in robotics, attention mechanisms have been extensively explored, aiming to model perception in agents also for socially interactive tasks~\cite{breazeal1999context}. Specifically, bio-inspired saliency approaches have shown effectiveness in tasks such as motion detection and exploratory gaze behavior~\cite{schauerte2011multimodal,ruesch2008multimodal}, including fast camera orientation towards human faces~\cite{butko2008visual} and UAV obstacle avoidance guided by saliency cues~\cite{ma2018multi}.

Attention mechanisms distribute processing among different feature extractions, combining multiple conspicuity maps to produce a final saliency map that guides robot head movements in real-time~\cite{ude2005distributed}. In these models, spatially localised stimuli compete for attention using mechanisms like \ac{WTA}. Similar attention approaches have been applied, integrating cues from vision, audition, and haptics~\cite{vijayakumar2001overt}. 
Other implementations focus their effort on introducing depth perception~\cite{kawabata1986attention, bruce2005attentional, pasquale2016enabling}. Motion also plays a significant role in attention mechanisms, essential for tracking objects or avoiding obstacles~\cite{thompson2012attention, li2019motion}.
The resulting saliency map can serve multiple purposes, including guiding selective visual attention~\cite{minut2001reinforcement}, facilitating object recognition~\cite{rutishauser2004bottom}, and enabling object tracking~\cite{mnih2014recurrent}. These applications demonstrate the utility of saliency-based models in identifying \acp{ROI} for subsequent localised processing.


These studies represent valuable efforts in developing attention systems to enable agents to engage in visual exploration of their environment. However, despite their emphasis on minimising computations by focusing solely on the highly resolved portion of the visual field, neither of these studies addresses two significant constraints relevant in real-time applications: latency and energy consumption. One potential solution involves exploring a fully bio-inspired pipeline to bridge the gap between bio-inspired software and hardware, thereby maximising the potential to reduce the computational burden and latency of the system by exploiting a sparse and event-based visual representation of the scene, based on event cameras. 

Event cameras~\cite{DVSsurvey} operate more akin to biological eyes than frame-based cameras. Instead of sequentially scanning each pixel to measure incident light levels, event cameras have independent pixels that generate spikes when the incident light surpasses a threshold. These ``pixel spikes" resemble the action potentials transmitted from the retina to the brain. The output of event cameras is asynchronous, sparse, and occurs only when there is a contrast between dark and light regions in the scene, detected as an illumination change for each pixel over time, essentially functioning as a dynamic edge extractor. 

Mechanisms of selective attention have been significantly advanced through the use of neuromorphic circuits coupled with event cameras~\cite{bartolozzi2009selective}. These mechanisms have been successfully implemented in humanoid robots equipped with neuromorphic cameras. This enables the robot to focus on the most salient regions of a scene~\cite{d2022event}.

Saliency-based models~\cite{itti1998model} typically involve three main stages: feature extraction, computation of individual feature maps, and integration and/or competition mechanisms to produce a final saliency map~\cite{zhao2013learning}. 
Additional recent works demonstrate the power of recurrent neural networks inspired by human sequential attention~\cite{mnih2014recurrent}, saliency-based noise mitigation with center bias incorporation~\cite{tong2015salient}, and deep attention models predicting human fixations~\cite{kummerer2017understanding}. 


Recent event-driven, saliency-based approaches have investigated, incorporating Gestalt principles~\cite{kohler1967gestalt} to identify salient regions likely to contain objects using cues such as intensity and depth~\cite{iacono2019proto}. The integration of neuromorphic sensing and computing, through event-based sensors and Spiking Neural Networks (\acp{SNN}), enables efficient, parallel processing of sparse visual data, inspired by the brain’s computational principles~\cite{gehrig2020eventbasedangularvelocityregression}. These methods have demonstrated effectiveness in real-time applications~\cite{d2022event}.
Building on this, a fully \ac{SNN}-based implementation of event-driven saliency has been deployed on neuromorphic hardware such as SpiNNaker for real-time performance~\cite{d2022event}. Furthermore, the incorporation of recurrent connections in the visual attention model~\cite{iacono2019proto} has been shown to improve object presence estimation and enable biologically plausible segmentation~\cite{d2024event}.

An important component often overlooked in these approaches is the downscaling of visual input, which could substantially reduce processing demands and latency in real-time scenarios~\cite{gruel2022event}. Previous studies have looked at identifying \acp{ROI} from  a downsampled visual field, however, these focused on selecting fixed portions of the visual field~\cite{gruel2023stakes}.

The downscaling aspect is becoming more relevant in modern applications. Recent event cameras~\cite{Prophesee_sensor} can achieve high spatial resolutions, enabling more accuracy in computer vision tasks, however this results in generating streams of events which are too intensive to be processed by real-time embedded systems. 
The EU NimbleAI project~\cite{NimbleDATE} aims to address this issue, by developing an integrated sensing and processing architecture able to operate on and control multiple input resolutions and to select only relevant information from the visual field, while discarding what is redundant.
This work, by extending the \ac{SNN} saliency-based model~\cite{angelo2025wandering,d2022event}, proposes an early perception selective attention architecture able to identify ROIs, with the aim of controlling multi-resolution event cameras~\cite{NimbleDATE}, and reducing down-stream processing.

This work builds on top of the neurmorphic, event-based, proto-object saliency model~\cite{d2022event}, adding multi-scale downscaling to further reduce the computational load and system latency. Key contributions include:
\begin{itemize}
    \item The final learning-free implementation operates at 1 ms latency, an order of magnitude faster than previous studies~\cite{AutomotiveDataset}, leveraging a lightweight shallow SNN with only two layers.
    \item A key extension of this work is its application to challenging real-world scenarios, such as automotive environments, where we demonstrate the model achieves 70\% accuracy in detecting salient objects, and ignoring clutter and noise.
    \item We investigate the role of input downscaling in the process of \ac{ROI} detection. We show that our model is robust to input degradation, achieving comparable performance with increasing downscaling factors, and is able to identify moving, rather than static, \acp{ROI} operating at 1 ms time resolution. This results in a significant reduction of processed data, over $19\times$ on the considered automotive use case.
\end{itemize}

The paper is structured into the following sections. Section \ref{sec:methods} presents an overview of the proposed pipeline and the methods used for this work, including an analysis of the potential benefits of such approach in terms of events reduction. Section \ref{sec:results} presents the experimental results. Finally, Section \ref{sec:conclusion} summarises the contributions and draws conclusions.


\section{Methods}
\label{sec:methods}

\begin{figure}
    \includegraphics[width=\linewidth]{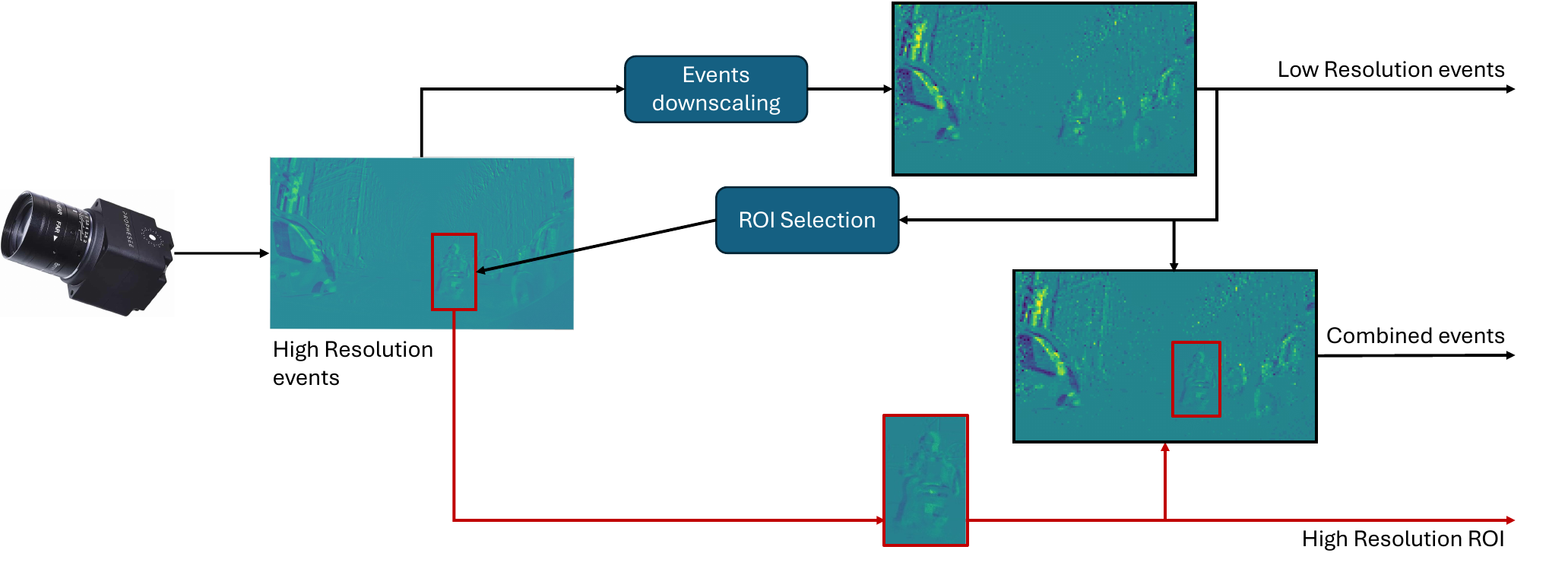}
    \caption{Proposed pipeline for multi-resolution \ac{ROI} detection. Visual events are downscaled, and low-resolution events are used to select relevant \acp{ROI}. Our proposed pipeline produces separate event streams, allowing for processing of low resolution and \ac{ROI} events separated, or in conjunction.}
    \label{fig:pipeline}
\end{figure}

Exploring the concept and potential of foveated sensing, as outlined in Section~\ref{sec:introduction}, requires sensing at multiple resolutions, and using this reduced resolution data to predict where in the visual field high-resolution sensing resource should be focused. While ideally this research would be performed on a multi-resolution hardware sensor, due to availability, this work explores a simulated sensor in order to observe the effect of resolution on \ac{ROI} detection. The proposed pipeline is shown in Figure~\ref{fig:pipeline}, demonstrating high-resolution sensor input on the left-hand side, and \ac{ROI} output on the bottom. 
Two distinct steps are required to enable the proposed pipeline: downscaling, and \ac{ROI} Selection. In this work, we focus on event data, and therefore downscale data produced using neuromorphic cameras (see Section~\ref{sec:downscaling} for algorithm details, and Section~\ref{sec:characterisation} for characterisation on common neuromorphic datasets). This downscaled data then forms the input to \ac{ROI} selection models, which are able to predict where features of interest are occuring in the data stream. A number of techniques could be used for \ac{ROI} detection, including trained classification and object-detection type networks. However, in this work, the focus is on low-latency \ac{ROI} detection from generic event streams, and therefore a bottom-up saliency-based attention model is selected (see Section~\ref{sec:model}). It's noted that for specific applications, alternative \ac{ROI} selection algorithms may yield improved performance, however the overall pipeline of Figure~\ref{fig:pipeline} remains the same, with the goal to compress the incoming data stream, by sensing the general field of view in low resolution, and sensing only \acp{ROI} in high-resolution. 

\subsection{Downscaling}
\label{sec:downscaling}

\begin{figure}
\centering
\begin{subfigure}{.5\textwidth}
  \centering
  \includegraphics[width=0.9\linewidth]{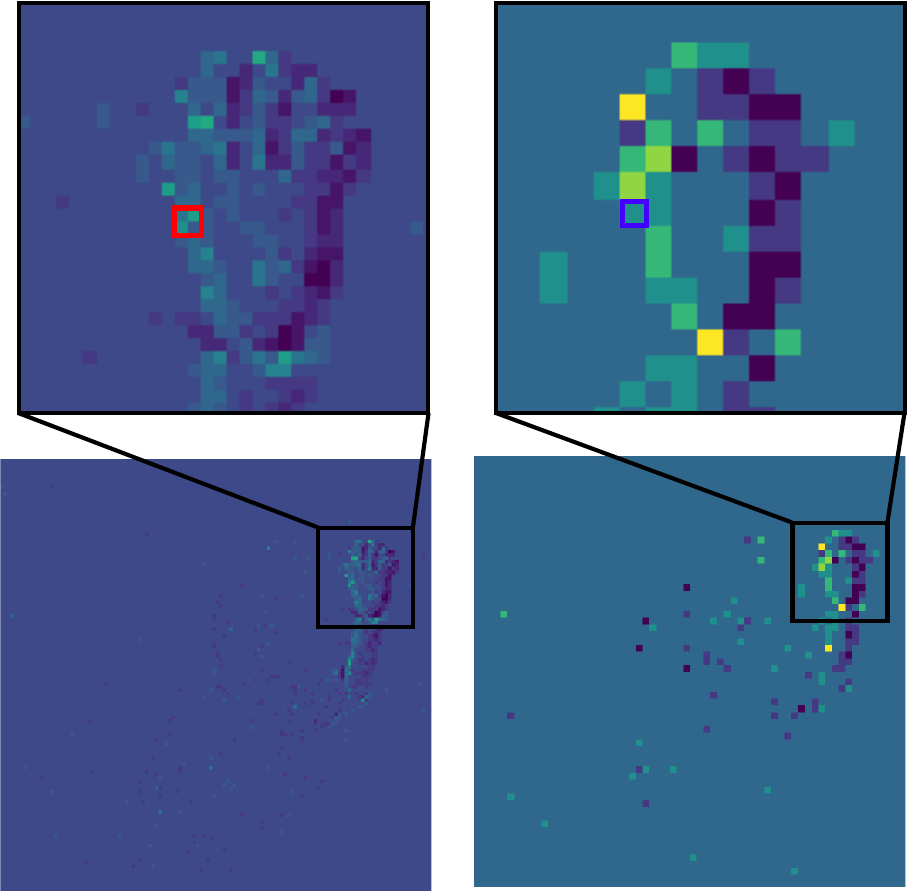}
  \caption{}
  \label{fig:hand}
\end{subfigure}%
\begin{subfigure}{.5\textwidth}
  \centering
  \includegraphics[width=0.9\linewidth]{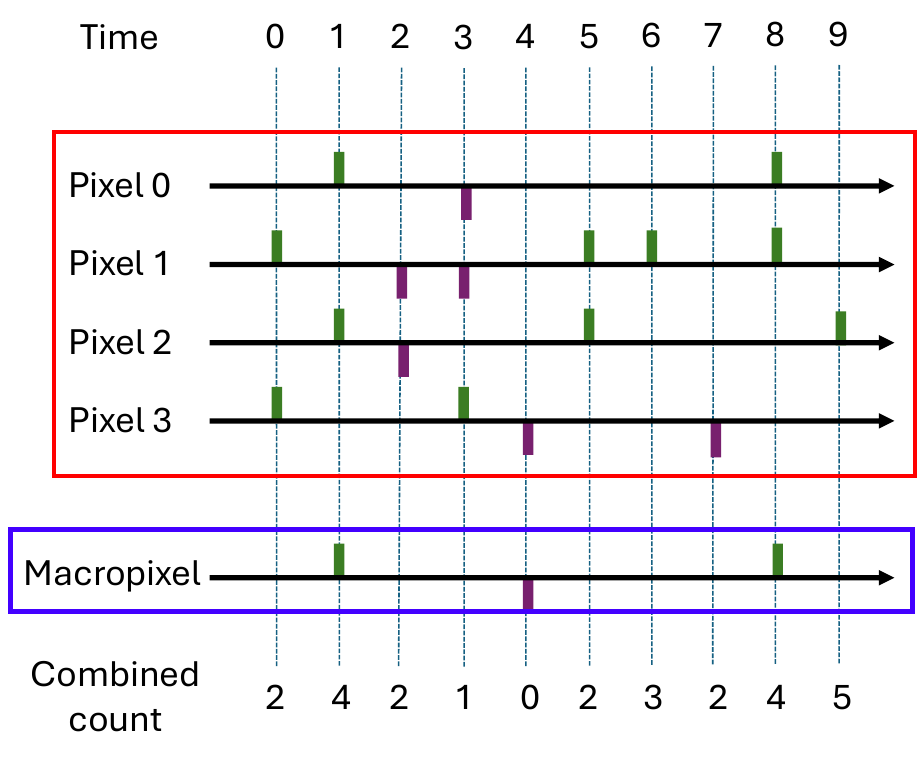}
  \caption{}
  \label{fig:times}
\end{subfigure}%
  \caption{Downscaling event resolution for a sample representing a subject waving their hand, extracted from the DvsGesture dataset~\cite{DVSGesture}, when applying a $4\times$ scaling factor and log luminance reconstruction based on event count. The left side (a) shows first a frame (accumulated over 10 ms) with events represented at scale, and then the same frame downscaled. The two insets show the \ac{ROI} enlarged, with pixels combined into a \emph{Macropixel} as highlighted by the red and blue squares. The right side (b) presents the events generated by the pixels in the region in red in the fullscale frame (clockwise order, from top left) over the 10 ms of execution, and how these are downscaled according to the combined threshold into the blue pixel in the downscaled frame.}
  \label{fig:downscaling}
\end{figure}

Spatial downscaling techniques reduce event throughput of event cameras and facilitate real-time processing in embedded systems~\cite{gruel2022event, gruel2023performance}.
While there are multiple techniques which can be applied, most exploit the concept of \emph{Macropixels}. A \emph{Macropixel} is obtained by combining the events from multiple adjacent pixels from the sensor grid in a square fashion, as shown in red in Figure~\ref{fig:hand} left. The downscaled output consists of an array of \emph{Macropixels}, as shown in Figure~\ref{fig:hand} right, generating events in a similar way to the original pixel. The downscaling factor corresponds to the number of original pixels belonging to each \emph{Macropixel} (e.g. Figure~\ref{fig:hand} right, uses a $2\times2$ \emph{Macropixel}, representing a $4\times$ downscaling of Figure~\ref{fig:hand} left). 
Downscaling techniques leveraging \emph{Macropixels} differ in the way the \emph{Macropixels} are generated and how they produce events.
The most effective event-based downscaling approaches~\cite{gruel2022event} aim at reconstructing the log luminance of a \emph{Macropixel} as the average of those pixels in the original sensor array which are part of the \emph{Macropixel}. The work presented in this paper builds on this idea, extending on previous work \cite{gruel2022event}, by estimating the mean log luminance of each \emph{Macropixel} through a normalised event count of the pixels belonging to it. The mechanism is illustrated in Figure~\ref{fig:times}, where the downscaled \emph{Macropixel} composed of 4 pixels ($2\times2$), from the events represented in Figure~\ref{fig:hand}, in the full-scale sensor grid is shown.
The output events of the pixels belonging to the \emph{Macropixel} are accumulated and then checked against a contrast threshold to determine whether an event is generated. Downscaled events are generated only when the entire threshold interval is covered (in analogy to what happens for single pixels in the original sensor array), therefore the slope is tracked, providing information on when an extremum happens (an example of this is shown in Figure~\ref{fig:times}, where the threshold for an OFF event is crossed at time 2, but an event is only generated at time 4, i.e. when the full threshold interval is covered and the threshold crossed).

This approach is useful for embedded real-world applications, as it allows for generating downscaled events in real time, while streamed from a sensor, as it does not require knowing when future events on the pixels involved in the downscaling process will be generated~\cite{gruel2022event}.

\subsection{Characterisation}
\label{sec:characterisation}
In this section we show the advantages of employing multi-resolution event processing in terms of output event reduction on datasets commonly used in the field. The analysis was performed on both the DvsGesture dataset~\cite{DVSGesture} representing different human gestures and the 1 Mpx Prophesee Automotive dataset~\cite{AutomotiveDataset} for dynamic driving environments.
Differently sized \emph{Macropixels} were generated, yielding different downscaling factors, with dynamic \acp{ROI} on the most active area of the visual field in the case of the DvsGesture dataset, and matching the provided bounding boxes for the Automotive dataset.

For the DvsGesture dataset, downscaling using \emph{Macropixels} of dimensions $2\times2$, $4\times4$, and $8\times8$ pixels was evaluated, corresponding to scaling factors of $4\times$, $16\times$, and $64\times$, respectively. The $8\times8$ macropixel case represents an extreme scenario, as the resulting visual field is reduced to a $16\times16$ pixel array. At such high downscaling levels, the averaging used in the log-luminance reconstruction introduces excessive noise, particularly given the relatively low original resolution of the DvsGesture recordings ($128\times128$ pixels). Consequently, higher downscaling factors are excluded from our analysis.
For the Automotive dataset, downscaling was performed using \emph{Macropixels} of sizes $2\times2$, $4\times4$, $8\times8$, and $16\times16$.

To evaluate the reduction in event count, in this experiment, we computed the total number of events generated by the sensor at full resolution and compared it with the corresponding number of events after downscaling (see Table~\ref{tab:downscaling_results}). The downscaled event counts include background activity, which represent the total number of events recorded in the downscaled visual field, excluding those within the ROIs. Events occurring within the \acp{ROI} are reported separately to analyse the significant difference in event count from the background activity. The total event reduction factor is also reported for each donwscaling. This is calculated on the total event count, i.e. including both low resolution and ROI events.
Additionally, Table~\ref{tab:downscaling_results} reports the temporal density, calculated over the number of events generated per millisecond; this metric is particularly relevant for real-time applications involving Spiking Neural Networks (\acp{SNN}), which frequently operate on neuromorphic hardware designed to emulate neural processing at millisecond-level temporal resolution~\cite{Real-time, parallelization}. 
Following previous studies~\cite{gruel2022event, gruel2023performance}, the temporal density of the event stream  is useful to quantify the average activation probability of pixels across the full sensor array. This is computed according to Equation~\ref{eq:temp_density}.

\begin{equation}
    \label{eq:temp_density}
    D = \frac{\sum_{x,y} P_{x,y}}{s}
\end{equation}

Where $P_{x,y}$ is the activation probability of each pixel in the considered time range, and $s$ is the sensor size.
$P_{x,y}$ can be obtained from Equation~\ref{eq:activ_prob}, where $e_{x,y}(t)$ are the events generated by the pixel of coordinates $(x, y)$ at time $t$, and $\Delta t$ is the time interval.

\begin{equation}
\label{eq:activ_prob}
    P_{x, y} = \frac{\sum_{t} e_{x,y}(t)}{\Delta t}
\end{equation}

The temporal density shows the distribution of events over the sensor, linking pixel activation to the amount of transmitted information. A high density corresponds to event data with high activation~\cite{gruel2023performance}.

\begin{table*}
    \centering
    \begin{tabular}{>{\centering\arraybackslash}p{0.13\linewidth}|>{\centering\arraybackslash}p{0.13\linewidth}|>{\centering\arraybackslash}p{0.13\linewidth}|>{\centering\arraybackslash}p{0.13\linewidth}|>{\centering\arraybackslash}p{0.13\linewidth}|>{\centering\arraybackslash}p{0.13\linewidth}}
    \hline
    \multicolumn{6}{c}{DVSGesture Dataset} \\
    \hline
    Downscaling & Combined Event Count & Background Event Count (LR) & \ac{ROI} Event Count (HR) & Total Event Reduction Factor & Temporal Density \\
    \hline
    Full scale & 6.89 $\times 10^5$ & N/A & N/A & N/A &  6.205$\times 10^{-6}$\\
    2$\times$2 & 4.69$\times 10^5$ & 3.77$\times 10^4$ & 4.31$\times 10^5$ & 1.47$\times$ & 4.038$\times 10^{-6}$\\
    4$\times$4 & 3.83$\times 10^5$ & 9.54$\times 10^3$ & 3.74$\times 10^5$ & 1.80$\times$ & 3.262$\times 10^{-6}$\\
    8$\times$8 & 2.12$\times 10^5$ & 2.20$\times 10^3$ & 2.09$\times 10^5$ & 3.25$\times$ & 1.797$\times 10^{-6}$\\
    \hline
    \multicolumn{6}{c}{Automotive Dataset} \\
    \hline
    Downscaling & Combined Event Count & Background Event Count (LR) & \ac{ROI} Event Count (HR) & Total Event Reduction Factor & Temporal Density \\
    \hline
    Full scale & 2.98$\times 10^8$ & N/A & N/A & N/A & 3.197$\times 10^{-5}$\\
    2$\times$2 & 5.98$\times 10^7$ & 4.51$\times 10^7$ & 1.47$\times 10^7$ & 4.98$\times$ & 6.131$\times 10^{-6}$\\
    4$\times$4 & 2.55$\times 10^7$ & 1.07$\times 10^7$ & 1.47$\times 10^7$ & 11.69$\times$ & 2.575$\times 10^{-6}$\\
    8$\times$8 & 1.72$\times 10^7$ & 2.44$\times 10^6$ & 1.47$\times 10^7$ & 17.32$\times$ & 1.731$\times 10^{-6}$\\
    16$\times$16 & 1.52$\times 10^7$ & 4.93$\times 10^5$ & 1.47$\times 10^7$ & 19.60$\times$ & 1.533$\times 10^{-6}$\\
    \hline
    \end{tabular}
    \caption{Execution results are reported for both the DvsGesture and Automotive datasets, showing the downscaling, the combined event count (HR+LR), the background events (LR), the \ac{ROI} events (HR), total events reduction and the temporal density. The evaluated time intervals are $6\mathrm{s}$ for the DvsGesture dataset (approximately the duration of one hand gesture), and $\sim10\mathrm{s}$ for the Automotive dataset. These values are obtained as average from execution on both the entire datasets.}
    \label{tab:downscaling_results}
\end{table*}

All listed metrics were measured for all applied downscaling factors, with results of all measurements, including event counts and temporal density, presented in Table~\ref{tab:downscaling_results}.
The time intervals used to evaluate the data presented in Table~\ref{tab:downscaling_results} are set to the average duration of one hand gesture ($\approx6\,\mathrm{s}$~\cite{DVSGesture}) for the DvsGesture dataset, and $\sim10\,\mathrm{s}$ of recording for the Automotive dataset, and are obtained from execution on the entire dataset for both the cases. For both the experiments, the combined event count decreases, as expected, with increasing downscaling factors, showing up to a $3.25\times$ reduction for the DvsGesture dataset, achieved with an $8\times8$ downscaling; and $19.6\times$ for the Automotive dataset with a $16\times16$ downscaling. The temporal density exhibits a consistent decrease across both experiments, highlighting the effectiveness of the approach in reducing data volume.
Once the system selects the \ac{ROI}, the original non-downscaled events from the event camera pixels are retrieved at full resolution (\ac{ROI} Events). In most cases, the number of events in the \ac{ROI} is comparable to the total number of downscaled events in the background. For high scaling factors ($16\times16$ for the Automotive dataset), the number of events in the \acp{ROI} is greater than the number of events in background (low resolution events).

\begin{figure}
  \includegraphics[width=\linewidth]{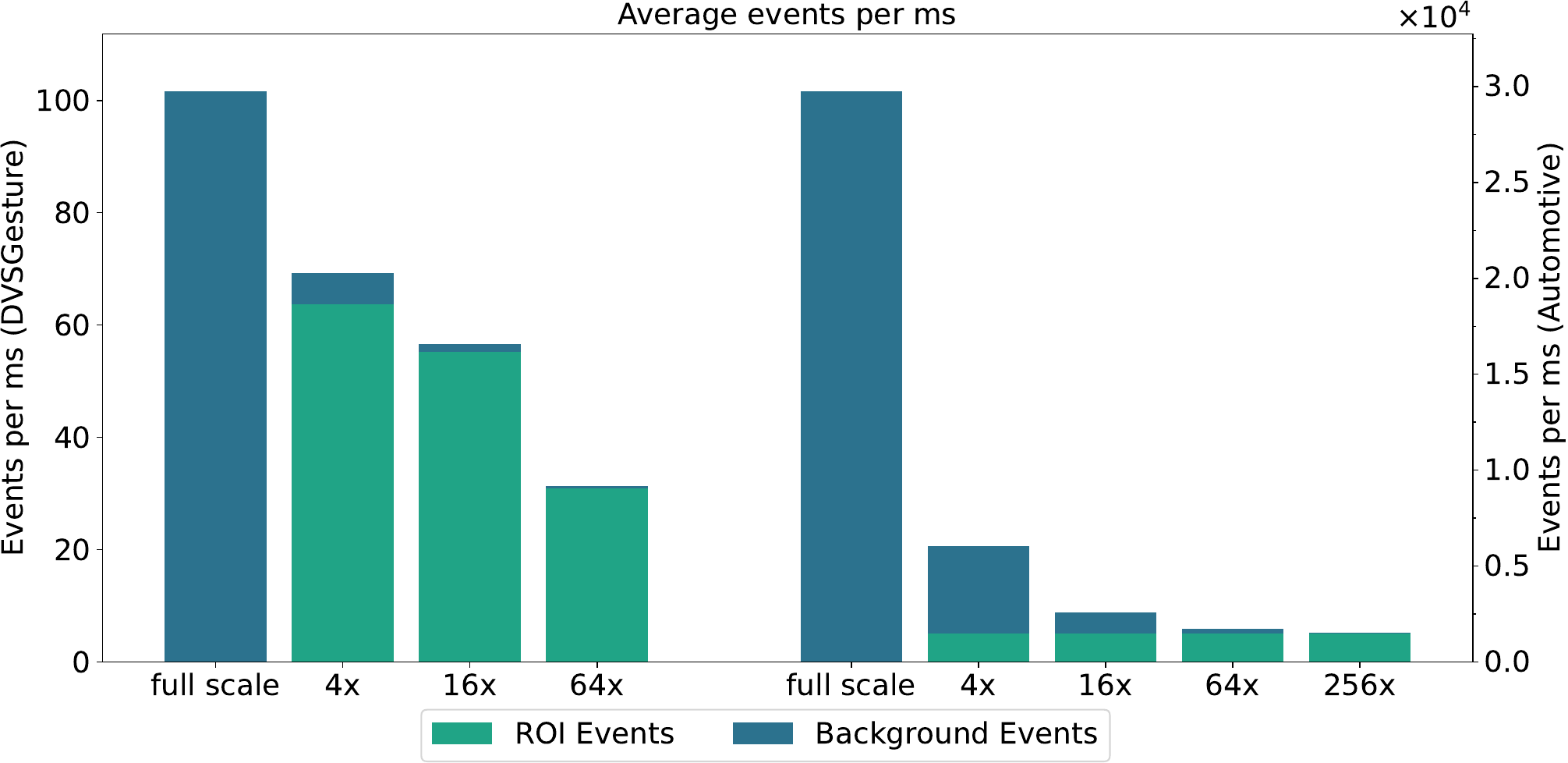}
  \caption{Histogram representing the generated average events per ms for the DVSGesture (4 bars on left) and the Automotive ( 5 bars on the right) datasets, with increasing downscaling factor (\emph{Macropixel} size). The proportion of events in the \ac{ROI} and in background is shown in the form of stacked bars.}
  \label{fig:evesms}
\end{figure}

Figure~\ref{fig:evesms} shows histograms with the number of generated events for each downsampling per millisecond, counted and averaged over the whole set, for both the DVSGesture (four bars on the left) and Automotive (the five bars on the right) datasets. Each bar shows the number of events that fall in the Low-Resolution (LR) Background area (dark blue) and those that fall in the High-Resolution (HR) \ac{ROI} area (light blue). These values serve as useful indicators for estimating input requirements in resource and time constrained applications, such as real-time \acp{SNN} inference on embedded neuromorphic hardware~\cite{Real-time}.
The generated events per millisecond decrease of  $\approx70\%$ with increasing downscaling factor for the DvsGesture dataset (up to $8\times8$ \emph{Macropixel}) , and decrease of about 95\% with a $16\times16$ downscaling for the Automotive dataset. This results in a reduction from $\sim$101 events per millisecond to $\sim$31 for the DVSGesture dataset, and from $\sim$29k to $\sim$1.5k for the Automotive case. With a downscaling factor reaching $256\times$ ($16\times16$ \emph{Macropixel}), the efficacy of the approach starts to diminish, since the number of \ac{ROI} events dominates the total event count. This is confirmed by the temporal density for the same case presented in Table~\ref{tab:downscaling_results}. However, this did not impact the visual consistency, and further downscaling factors could still be applied.
Table~\ref{tab:downscaling_results} shows that the reduction in temporal density is higher for the 1~Mpx sensor, compared to the DVS128 sensor, hence better compressing the per pixel activation.

\subsection{Saliency-Based \ac{ROI} Estimation}
\label{sec:model}

Building on top of the benefits presented in section~\ref{sec:characterisation} in terms of event reduction, this work focuses on the development of biologically-inspired saliency-based methods as a mechanism to guide \ac{ROI} selection. This builds on the proto-object model initially developed by~\cite{Russell2014a} for frame-based cameras and later adapted for event-driven cameras~\cite{iacono2019proto} and modified as \ac{SNN}~\cite{d2022event}. Its architecture includes two key layers: Border Ownership and the Grouping Pyramid. These layers incorporate principles of Gestalt psychology~\cite{kohler1967gestalt}, including continuity and figure-ground organisation, to process closed object contours at multiple scales, ensuring scale invariance.

\begin{figure*}
    \centering
    \includegraphics[width=\linewidth]{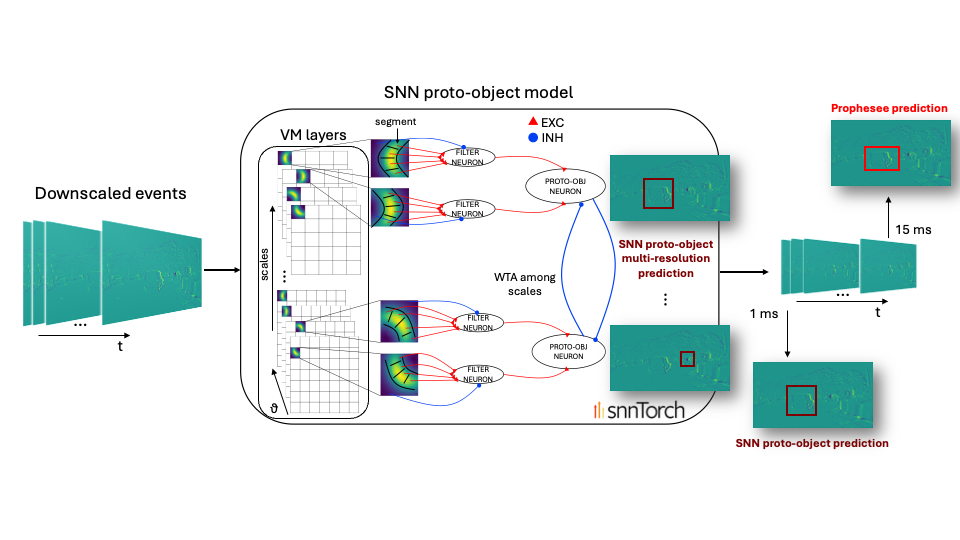}
    \caption{Network structure. Events binned at 1 ms time resolution are filtered by \ac{VM} filters of different sizes and orientations at different scales, feeding into the filter neurons layer. Neighbouring filters with opposite convexities and the same size are then combined into the proto-object layer (grouping layer). Lateral inhibition in the proto-object layer selects a single \ac{ROI}.}
    \label{fig:architecture}
\end{figure*}

The implementation presented here is inspired by the \ac{SNN} saliency-based visual attention model running on the SpiNNaker neuromorphic hardware~\cite{d2022event}. The key contribution of our work lies in the model's ability to track salient regions at different scales and generate adaptable \acp{ROI} over the detected proto-objects.
The first layer of the saliency-based visual attention model, the Border Ownership Pyramid, takes advantage of von Mises (\ac{VM}) filters, depicted in Figure~\ref{fig:architecture}, to detect close contours (See Eq.~\ref{eq:VM}):

\begin{equation}\label{eq:VM} 
VM_{\theta} (x,y) = \frac{\exp(\rho \cdot R_0 \cdot \cos(atan2(-y, x) - \theta)}{I_0(\sqrt{x^2 + y^2} - R_0)} 
\end{equation}

Where $x$ and $y$ represent the coordinates of the kernel centered at the origin, $R_0$ denotes the radius of the filter, $\rho$ controls the arc length of the active pixels within the kernel, influencing its convexity, $\theta$ specifies the orientation, and $I_0$ refers to the modified Bessel Function of the first kind.
The initial layer of the model employs $m \times l$ \ac{VM} filters, with different sizes (in the following $m$=3) and a specific orientation (in the following $l$=8) positioned on the input layer with $stride=1$. Each \ac{VM} filter possesses a distinct receptive field, such that each incoming event activates a specific pixel within one of these filters. 
The kernels are connected to a single \textit{Filter Neuron} via excitatory connections. Additionally, the \textit{Filter Neuron} receives inhibitory input from the background of the \ac{VM} kernel, enhancing the selectivity of the receptive field to curved shapes. This stage of the model represents the Border Ownership cells.  

\textit{Filter Neurons} of opposite orientations of the \ac{VM} kernels are connected with each other to perceptually group the information resembling the Grouping cells layer, referred to as \textit{Proto-object Neurons}, correlating to the presence of an object of a given size. These contribute to the formation of the saliency map. The filters are arranged into four rotational pairs (8 orientations), with each pair rotated 180 degrees relative to its counterpart. The outputs from all spatial scales and rotational pairs are then integrated to generate the final saliency map. A \ac{ROI} is subsequently defined around the point of maximum saliency, by means of a \ac{WTA} approach, where the most salient proto-object inhibits all the others, resulting in selection of a single \ac{ROI}.
The output of the network, therefore, provides the centre coordinates and bounding box of the \ac{ROI}, as described in Section~\ref{sec:model}.

\section{Experiments \& Results}
\label{sec:results}
 
The presented model is tested against 4 different downscaling factors (4$\times$, 16$\times$, 64$\times$, 256$\times$), mirroring the analysis performed in Section~\ref{sec:characterisation}, to show its efficacy on multiple resolutions. For each scaling factor, 3 different sizes of \ac{VM} filters are applied alongside the scaling factor of the visual field, with the filter overlap kept at 99\%. 
In order to quantitatively estimate the accuracy of the presented approach, the accuracy of the detected \acp{ROI} is checked against dataset's annotations (bounding boxes) as ground truth, and the overlap between the detected \acp{ROI} and the bounding boxes is measured. Therefore only annotated datasets are suitable for such analysis. The overlap between \acp{ROI} and bounding boxes is obtained by calculating the percentage of the ROI, in terms of pixels, which lies within a single bounding box. 
Here we report the total average overlap of all the \acp{ROI} to the bounding boxes.

As an additional metric, we show the \ac{IoU} of the detected ROIs, compared to the dataset's bounding boxes. 
The detected \acp{ROI} from the presented implementation are restricted to being square and assume pre-defined fixed-size dimensions, while the ground truth bounding boxes vary in shape and size. For this reason,the potential maximum \ac{IoU}, obtained by aligning the top left corner of a \ac{ROI} and the relative bounding box, is calculated, and the real \ac{IoU} is divided by this value, obtaining a normalised \ac{IoU}. 
This returns a relative measure of the IoU with respect to the maximum value that could be obtained if the \ac{ROI} and the bounding box were aligned. 
The \ac{IoU} metric is calculated according to Equation~\ref{eq:Iou}.
\begin{equation}
\label{eq:Iou}
    IoU = \frac{|ROI \cap BB|}{|ROI \cup BB|}
\end{equation}
In equation~\ref{eq:Iou}, the term $ROI$ refers to the region of interest detected by our model, while $BB$ refers to the annotated bounding boxes from the dataset.

\subsection{The Prophesee Automotive Dataset}
The 1~Megapixel Prophesee Automotive Dataset~\cite{AutomotiveDataset} contains 14.6 hours of recordings taken from a car driving under different scenarios (city, highway, countryside, small villages and suburbs), under different light and weather conditions during daytime. The dataset is provided together with annotations, in the form of 25 million bounding boxes of cars, pedestrians, two-wheeled vehicles, trucks, buses, traffic signs and traffic lights with a time resolution of 16 ms. The data was recorded using a high resolution (1280 x 720) event camera~\cite{Prophesee_sensor}, paired with a frame-based RGB camera, providing support in obtaining the labels~\cite{AutomotiveDataset}. To date, this is the largest publicly available event-based dataset.

The results presented for this work are based on this dataset, as it represents a real-world use case exhibiting a wide range of classes, with a spatial resolution high enough to allow for exploration of a range of downscaling factors.
Other datasets were considered for this study, such as the SalMapIROS dataset~\cite{iacono2019proto} and the IBM DVSGesture dataset~\cite{DVSGesture}. However, these were discarded due to a lack of labels to compare the detected \acp{ROI} with and due to a reduced input spatial resolution, as highlighted in Section~\ref{sec:characterisation}. Furthermore, recordings from such datasets typically tend to be short and hand crafted, removing the real-world context that this study is seeking.

In order to best estimate the sizes for the detected \acp{ROI}, statistics analysis of the Automotive dataset's bounding boxes was performed. This analysis is presented in Table~\ref{tab:Dataset_stats}. The average and median values for all the bounding boxes are computed and are respectively $105\times126$ and $64\times84$. Alongside these figures, statistics for each class are provided. These are the average and median bounding box sizes, the number of occurrences relative to the total number of bounding boxes for a given class (given as a percentage), and the mean aspect ratio of the boxes for that class. As can be seen from Table~\ref{tab:Dataset_stats}, \emph{Cars} is the most prominent class (amounting to 49.11\% of the bounding boxes), followed by \emph{Pedestrian} (22.30\%), \emph{Traffic signs} (9.36\%), \emph{Traffic lights} (9.04\%), \emph{Trucks} (5.46\%), \emph{Two wheelers} (3.27\%) and \emph{Buses}(1.45\%).
Given these measurements, the following sizes were chosen for the \acp{ROI} to be selected by our model: $100\times100$, $120\times120$, $150\times150$. These take into account the total average size, but also the spread among classes. Larger sized \acp{ROI} were preferred to ensure the calculated performance scores do not yield higher results due to the identification of portions of bounding boxes, rather than full objects. 
The overlap scores are calculated in terms of \ac{ROI} pixels overlapping labelled objects pixels. Should the former be fully contained in the latter, this would result in a 100\% overlap. To avoid these cases, the \ac{ROI} sizes were selected favouring the higher bounding box sizes, with respect to the average.

Another important aspect is the average aspect ratio of the bounding boxes from various classes. This metric indicates the average shape of a bounding box from a given class, where a value of 1 corresponds to a perfectly squared bounding box. Since the detected \acp{ROI} are squared in shape, the network will be more sensitive to bounding boxes of this sort. Therefore, classes such as pedestrians (with an aspect ratio of 0.34), two wheelers (with 0.61) and traffic lights (with 0.70) are expected to yield lower detection accuracy.

\begin{table*}
    \centering
    \begin{tabular}{c|c|c|c|c}
    \hline
    \multicolumn{5}{c}{Prophesee Automotive Dataset Statistics} \\
    \hline
    Total Mean Bounding Box Size & \multicolumn{4}{c}{105$\times$126}\\
    \hline
    Total Median Bounding Box Size & \multicolumn{4}{c}{64$\times$84}\\
    \hline
    Class & Median BB Size & Mean BB Size & Percentage & Aspect Ratio\\
    \hline
    Cars & 89$\times$64 & 135$\times$114 & 49.11\% & 1.18\\
    Pedestrians & 58$\times$112 & 29$\times$84 & 22.30\% & 0.34\\
    Traffic signs & 25$\times$28 & 45$\times$38 & 9.36\% & 1.18\\
    Traffic lights & 21$\times$44 & 40$\times$57 & 9.04\% & 0.70\\
    Trucks & 100$\times$100 & 159$\times$182 & 5.46\% & 0.87\\
    Two wheelers & 64$\times$113& 101$\times$163 & 3.27\% & 0.61\\
    Buses & 146$\times$127 & 200$\times$216 & 1.45\% & 0.92\\
    \hline
    \end{tabular}
    \caption{Statistics measured from the Prophesee automotive dataset per class.}
    \label{tab:Dataset_stats}
\end{table*}

The model was run at 1~ms time resolution, predicting a \ac{ROI} every timestep. This is 16 times finer than the time resolution at which the Automotive dataset bounding boxes are provided. This also aligns with on-line requirements for future mapping on neuromorphic hardware. Results are presented in Table~\ref{tab:results}, evaluated from the whole Prophesee Automotive dataset (14.6 hours of recording) for each downscaling factor. 
For each downscaling factor, we report the fraction of \acp{ROI} overlapping with the corresponding bounding box, the average overlap with standard deviation, and the normalised \ac{IoU} average and standard deviation.

The number of \acp{ROI} overlapping with the bounding boxes of the dataset ranges from $\sim 71\%$ at full scale, to $\sim 57\%$ with a $256\times$ downscaling factor. The average overlap is above $50\%$ for all cases. These results show the robustness of the approach to various downscaling factors, achieving minimal degradation of performance even with aggressive downscaling factors, while reaching prediction with 1~ms time resolution. As described in Section~\ref{sec:characterisation}, a $256\times$ downscaling results in an event reduction of approximately $20\times$ compared to a full scale representation. The accuracy in \ac{ROI} detection using our method however only degrades by $\sim 14\%$, while maintaining a similar average overlap ($\sim 51\%$ against $\sim 59\%$). The most notable case is represented by the $4\times$ downscaling, which yields a $5\times$ reduction in events, and a $\sim 65\%$ match for the \acp{ROI} (only a $\sim 6\%$ reduction compared to the full scale case).

\begin{table*}
    \centering
    \begin{tabular}{c|c|c|c|c|c}
    \hline
    \multicolumn{6}{c}{Mean \ac{ROI} Overlap} \\
    \hline
    Downscaling & Overlapping \acp{ROI} & Mean \ac{ROI} Overlap & Stdev & Mean IoU & Stdev\\
    \hline
    Full scale &  70.81\% & 58.67\%& 15.91\%& 61.54\% & 14.88\%\\
    4x & 64.89\% & 55.67\%& 13.89\%& 59.82\% & 11.89\%\\
    16x & 62.82\% & 53.25\%& 7.54\%& 56.59\% & 6.28\%\\
    64x & 58.06\% & 52.84\%& 5.40\%& 57.09\% & 4.89\%\\
    256x & 56.93\% & 50.61\%& 11.22\% & 52.11\% & 7.21\%\\
    \hline
    \end{tabular}
    \caption{Mean accuracy values for \ac{ROI} overlap and relative IoU for the various downscaling factors}
    \label{tab:results}
\end{table*}

\begin{figure}
\centering
  \includegraphics[width=\linewidth]{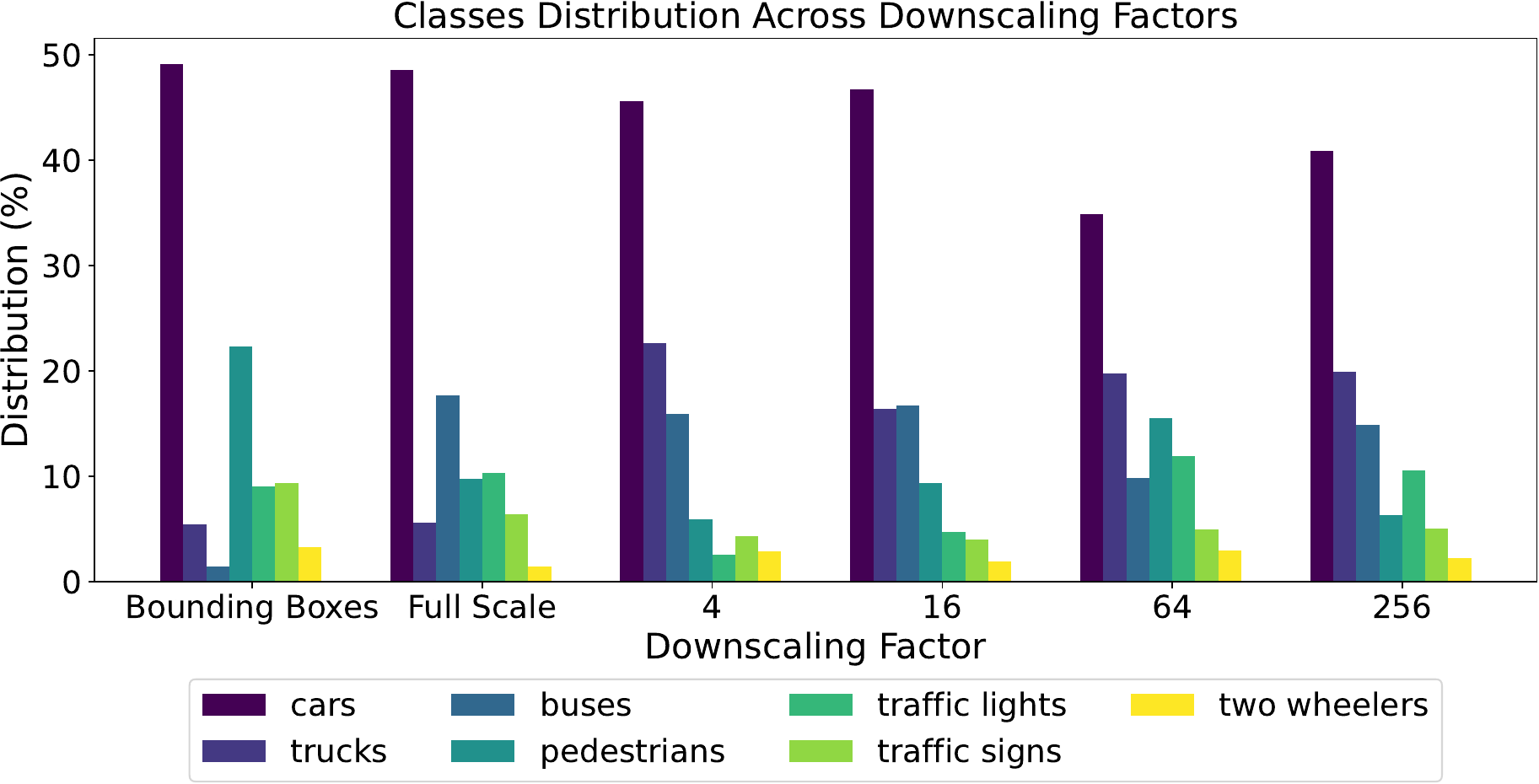}
  \caption{Distribution of \acp{ROI} per class across the various downscaling factors. Dataset bounding boxes values are on the left.}
  \label{fig:class_sensitivity}
\end{figure}

Figure~\ref{fig:class_sensitivity} shows the distribution of classes among the detected \acp{ROI} for each applied downscaling. For reference, the first provided case corresponds to the dataset bounding boxes, as shown in Table~\ref{tab:Dataset_stats}.
The model achieves similar class distibutions for all the downscaling factors. The \emph{Cars} class is the most prominent across different downscalings, being above $40\%$ for all cases (except for 64$\times$ downscaling). This is due to multiple factors: \emph{Cars} is by far the most common class, with $49\%$ of occurrences in the ground truth. This also means that the bounding boxes' mean size is mostly influenced by this value, contributing to the \ac{ROI} sizes chosen for the model. Furthermore, elements belonging to \emph{Cars} have on average an aspect ratio of 1.18, which makes them close to a square, therefore, more easily detectable by the network.
\emph{Pedestrians}, on the other hand, are not easily detected. This is shown by a large discrepancy between the bounding boxes dataset (at 22.3\%) and the detected \acp{ROI} belonging to this class (ranging from 5.95\% at 4$\times$ to 15\% at 64$\times$). This is mainly due to the nature of this class, which, on average, presents very rectangular elements, as evidenced by the low aspect ratio of 0.34. Despite this being the second most common class (at 22.30\%), only a few elements of this are detected across all the scaling. 
This is confirmed by the better performance in detecting \emph{trucks} and \emph{buses}, which again are approximately square (with aspect ratios of 0.87 and 0.92, respectively) and aligned with the bounding boxes in size. This becomes more evident with increasing downscalings, especially for high factors, where lower resolution causes classes to be less identifiable.

As highlighted in Section~\ref{sec:model}, the model looks to identify a single \ac{ROI} at a time, as the most salient portion of the visual field. In contrast, the annotations of the dataset often present multiple bounding boxes at the same time. For this reason, the class distribution presented in Figure~\ref{fig:class_sensitivity} exhibits higher values for some specific classes (e.g. \emph{busses} and \emph{trucks}), compared to the dataset bounding boxes. This is not to be attributed to false predictions, as the identified classes are only marked based on the overlap measurements, but to a different distribution of such classes, due to the nature of the task and the configuration of the model.

\begin{figure}[h]
\centering
  \includegraphics[width=\linewidth]{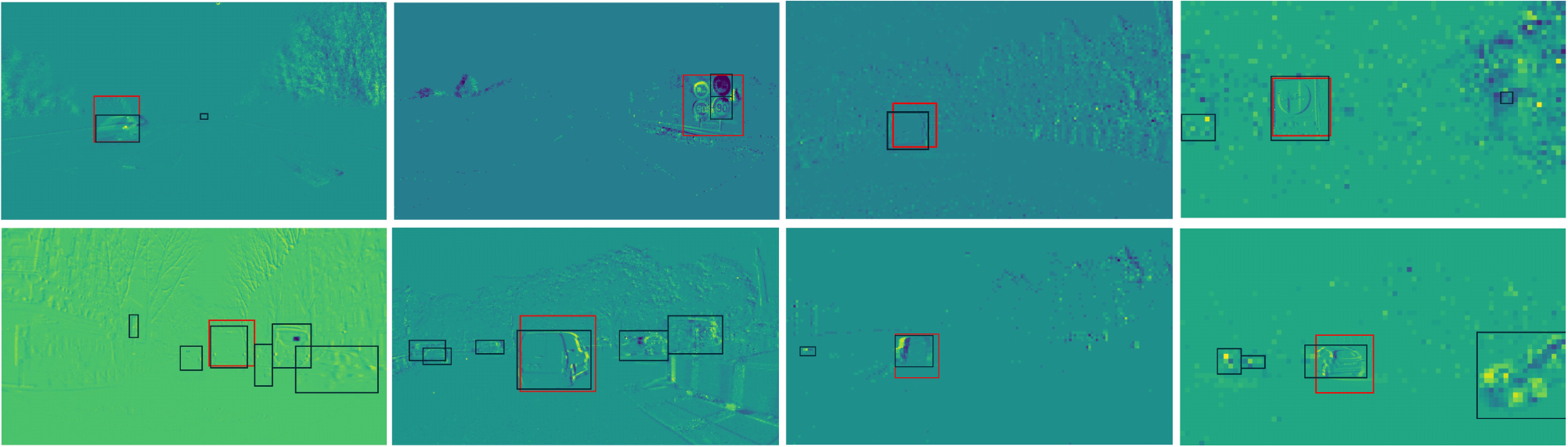}
  \caption{Examples of detected \acp{ROI} (in red) compared with the dataset bounding boxes (in black) at different downscaling factors (1 per column from left to right 4$\times$, 16$\times$, 64$\times$, 256$\times$). Frames are generated at 16~ms (bounding boxes time resolution) and events are displayed as heatmaps.}
  \label{fig:detected_rois}
\end{figure}

Finally, Figure~\ref{fig:detected_rois} shows examples of execution of the model. Eight separate cases are presented, randomly sampled from the 14.6 hours of recordings. Two cases per downscaling are reported. Each column represents one of the applied downscalings. Each frame is the result of event accumulation over the dataset bounding boxes period (i.e. 16 ms). Events are displayed as heatmaps, hence the variations in colour for the various cases. The red box corresponds to the \ac{ROI} detected from the model, the black boxes are the dataset's bounding boxes. For visual comparison, only the last identified \ac{ROI} in the presented time interval is displayed. Inside each detected ROI, events are presented at the original resolution, showing the multi-resolution effect achieved by the approach. Various classes are detected here, including \emph{Cars}, \emph{Trucks} and \emph{Traffic signs}. The presented samples show both good and bad overlap scores, where a good example is given by the top right case (256$\times$ downscaling), where a truck is identified with a high overlap score. A bad prediction is shown in the second column (16$\times$ downscaling), top frame, where, despite the \ac{ROI} containing the traffic signs, this is considerably larger than the bounding boxes, resulting in a very low overlap score, since this is calculated in terms of proportion of \ac{ROI} pixel overlapping with bounding box pixels.
All the presented cases in Figure~\ref{fig:detected_rois} show high event counts associated with trees in the background, and these are not influencing the network's predictions, demonstrating the capability of the model to ignore noise and clutter.

\section{Discussion}
\label{sec:conclusion}

We proposed an efficient event-based, high temporal resolution, learning-free, scalable, and general approach for \ac{ROI} detection. 
The system is a bottom-up model capable of extracting salient information at multiple resolutions from the visual scene at 1~ms temporal resolution. The system has been benchmarked against a real-world use-case, the Prophesee Automotive dataset, i.e. the largest event-based dataset available, showing an average of 62.7\% ROI detection accuracy, and 16$\times$ faster than the ground truth.

The first experiment shows the benefits of event downscaling in terms of data reduction, showing a close to exponential decrease in the amount of input events from the high resolution to the largest downscaling factor. 

The results show that our multi-scale approach is able to reduce information loss, caused by input event downscaling, by extracting high resolution \acp{ROI} from the visual field. While increasing downscaling factors introduces growing information distortion, our approach demonstrates robustness by achieving similar predictions across scales. This yields a more compact and efficient event representation with minimal loss of salient information. 

Previous studies \cite{angelo2025wandering} demonstrated that the detection of salient objects can reach an accuracy of 88.8\% in office scenarios and 89.8\% in challenging indoor and outdoor low-light conditions, as evaluated on the event-assisted low-light video object segmentation dataset \cite{li2024event}. Therefore, we compared our \acp{ROI} with the Automotive dataset bounding box annotations, achieving more than 70\% overlap of \acp{ROI} at full scale, with mean normalised \ac{IoU} values above 60\%, indicating strong alignment with the ground truth. Even under downscaling, the overlap remains consistently high, with values above 56\% for \acp{ROI} correspondence and mean normalised IoU above 50\%. This demonstrates that our training-free strategy estimates well the location of meaningful \acp{ROI}. Furthermore, our presented method is able to perform \ac{ROI} detection at 1 ms time resolution, achieving 1 order of magnitude higher resolution than ground truth lables provided with the Prophesee Automotive dataset. The results confirm that the model optimally detects cars, trucks, and buses within the objects in the dataset. 

The results presented here are well aligned with the objectives of the NimbleAI project~\cite{NimbleDATE}, targeting development of foveated vision systems. The presented model, due to its lightweight nature, can be deployed for real-time early perception and near-sensor selective attention, receiving as input low resolution events, as recorded by a multi-resolution \ac{DVS} sensor, and it is able to produce coordinates for \ac{ROI} selection, informing such a sensor where to focus higher resolution.
This work paves the way for the development of digitally foveated event cameras by providing the tools to explore event reduction strategies which are application agnostic. The development of a paired hardware accelerator, as part of the NimbleAI project, will allow for the deployment of such a model in/near the foveated sensor, allowing for real-time control. It also paves the way for multi-sensor systems, where low-resolution always-on sensors can be paired with higher resolution sensors, which do not have the power or data-storage budgets to record a wide field of view or continuously. The presented approach could be implemented on the low-resolution sensor to help guide data capture from the higher resolution sensor, creating an overall system with unique energy, latency and resolution characteristics. 

This strategy has the potential to be expanded and fine-tuned for more specialised tasks, by developing, for instance, top-down-based classifiers, which can inspect the selected \ac{ROI} looking for application-specific features. Such classifiers can then feed back to the bottom-up model described here, indicating how good a prediction is, and subsequently steer the network output towards other portions of the visual field, in case of a bad estimate, or maintain the focus on the selected area, in case of a good one.
This approach would allow for the development of more lightweight, therefore also faster, classifiers, which only need to focus on a ROI-sized input, rather than the entire full-resolution of the visual field, and are more specialised to best fit specific applications.
Extensions to this approach are also possible to detect multiple \acp{ROI} at once, by relaxing the network's output \ac{WTA} mechanism, to select the top $K$ most relevant objects. However, such a task has a different scope than what is explored in this work, and therefore is currently beyond our exploration. Future work could also include temporal integration as done by Chane et al.~\cite{chane2024eventbasedsaliency}, where the accumulation of the events and the subsequent saliency map is generated from multi-scale spatiotemporal volumes. However, this would need to be traded off against the low-latency capability demonstrated in this work. 

\section*{Author Contributions}
LP co-designed the selective attention model, developed the software implementation, ran the experiments and co-drafted the manuscript. GDA co-designed the model and co-drafted the manuscript. CB and OR supervised the research. All authors read, commented and approved the final manuscript.

\section*{Funding}
This research was supported through the NimbleAI project, funded via the Horizon Europe Research and Innovation programme (Grant Agreement 101070679), and UKRI under the UK government’s Horizon Europe funding guarantee (Grant Agreement 10039070); the Horizon Europe AIDA4Edge project (Grant Agreement 101160293), and under the UK government's Horizon Europe funding guarantee (10116105); and the EPSRC Edgy Organism project (EP/Y030133/1). GDA acknowledges the financial support from the European Union’s HORIZON-MSCA-2023-PF-01-01 research and innovation programme under the Marie Skłodowska-Curie grant agreement ENDEAVOR No 101149664.

\section*{Data Availability Statement}
The data that support the findings of this study is available at the following URL: https://github.com/NimbleAI-Manchester/NimbleAI.

\bibliography{bibliography}
\bibliographystyle{unsrt}

\end{document}